\documentclass[letterpaper]{article} 
\usepackage{aaai2027}  
\nocopyright
\usepackage[hyphens]{url}  
\usepackage{graphicx} 
\usepackage{natbib}  
\usepackage{caption} 
\usepackage{algorithm}
\usepackage{algorithmic}
\usepackage{amsmath}
\usepackage{amsfonts}
\usepackage{makecell}
\usepackage[most]{tcolorbox}
\usepackage[export]{adjustbox}
\usepackage{listings}
\DeclareCaptionStyle{ruled}{labelfont=normalfont,labelsep=colon,strut=off} 
\floatstyle{ruled}
\newfloat{listing}{tb}{lst}{}
\floatname{listing}{Listing}

\usepackage{booktabs}
\usepackage{pifont}
\usepackage{tabularx}
\definecolor{PromptBlue}{RGB}{78,164,211}
\definecolor{PromptBlueLight}{RGB}{225,239,248}

\providecommand{\pgopd}{PG-OPD}

\providecommand{\cmark}{%
  \textcolor{green!45!black}{\ding{51}}%
}
\providecommand{\xmark}{%
  \textcolor{red!70!black}{\ding{55}}%
}

\definecolor{QuestionBack}{RGB}{246,247,249}
\definecolor{QuestionFrame}{RGB}{110,118,129}
\definecolor{FullBack}{RGB}{253,242,242}
\definecolor{FullFrame}{RGB}{185,65,65}
\definecolor{PGBack}{RGB}{241,249,243}
\definecolor{PGFrame}{RGB}{58,135,82}

\title{Prefix-Guided On-Policy Distillation: Mining Golden Trajectories from Rollouts}

\author{
  Qingfei Zhao\textsuperscript{1,2} \quad
  Huan Song\textsuperscript{1} \quad
  Shuyu Tian\textsuperscript{1} \quad
  Jiawei Shao\textsuperscript{1} \quad
  Xuelong Li\textsuperscript{1}
  \\
  \textsuperscript{1}TeleAI \quad
  \textsuperscript{2}Shanghai Jiao Tong University
}
\affiliations{}
\begin{document}
\maketitle

\begin{abstract}
\begin{quote}
On-policy distillation (OPD) improves reasoning models by applying dense teacher supervision on student-sampled trajectories. However, scaling OPD to long-horizon reasoning exposes a reliability and efficiency problem: standard OPD assigns every candidate the same long rollout budget, even though some trajectories may quickly become weakly aligned with the teacher and provide less useful supervision. Prior analyses suggest that teacher--student
compatibility is important for OPD success, motivating early-prefix top-$k$
overlap as a proxy for continuation value. Continuing low-overlap trajectories
to the full rollout length may cause them to drift further from the teacher as
generation proceeds, increasing computational cost while providing limited
distillation benefit. To address this, we introduce \textbf{Prefix-Guided On-Policy Distillation (PG-OPD)}, a rollout-allocation framework that estimates continuation value from fixed-length prefixes. PG-OPD computes teacher--student top-$k$ overlap in an early probe window and allocates long rollouts only to high-overlap candidates, while stopping the rest at the prefix length. Across teacher--student combinations on AMC, AIME, and HMMT benchmarks, PG-OPD achieves up to a 4.80-point accuracy gain and up to a $2.46\times$ wall-clock speedup across configurations. These results demonstrate that using early-prefix compatibility to guide candidate pruning can improve both training efficiency and reasoning performance.

\end{quote}
\end{abstract}

\section{Introduction}

On-policy distillation (OPD) has emerged as a promising post-training paradigm for improving language-model reasoning. Conventional knowledge distillation transfers knowledge from a strong teacher to a smaller student by matching predictive distributions or learning from teacher-generated sequences \citep{hinton2015distilling, kim2016sequence, sanh2019distilbert, jiao2020tinybert, wang2020minilm}. In reasoning tasks, however, sequence-level off-policy distillation typically supervises the student on teacher-generated trajectories, which can create a train--inference mismatch: during training, the student conditions on teacher-generated prefixes, whereas at inference time it must continue from its own generations \citep{bengio2015scheduled}. OPD mitigates this mismatch by sampling trajectories from the current student policy and evaluating teacher supervision on the resulting student-visited states \citep{gu2024minillm, agarwal2024gkd}. Because supervision is available throughout the generated trajectory, OPD provides denser token-level learning signals than objectives based only on final-answer rewards, making it particularly appealing for long-horizon reasoning tasks that require
sustained multi-step generation \citep{yang2025qwen3, deepseekai2025deepseekr1, yu2025dapo}.

\begin{figure}[t]
\centering
\includegraphics[width=\columnwidth]{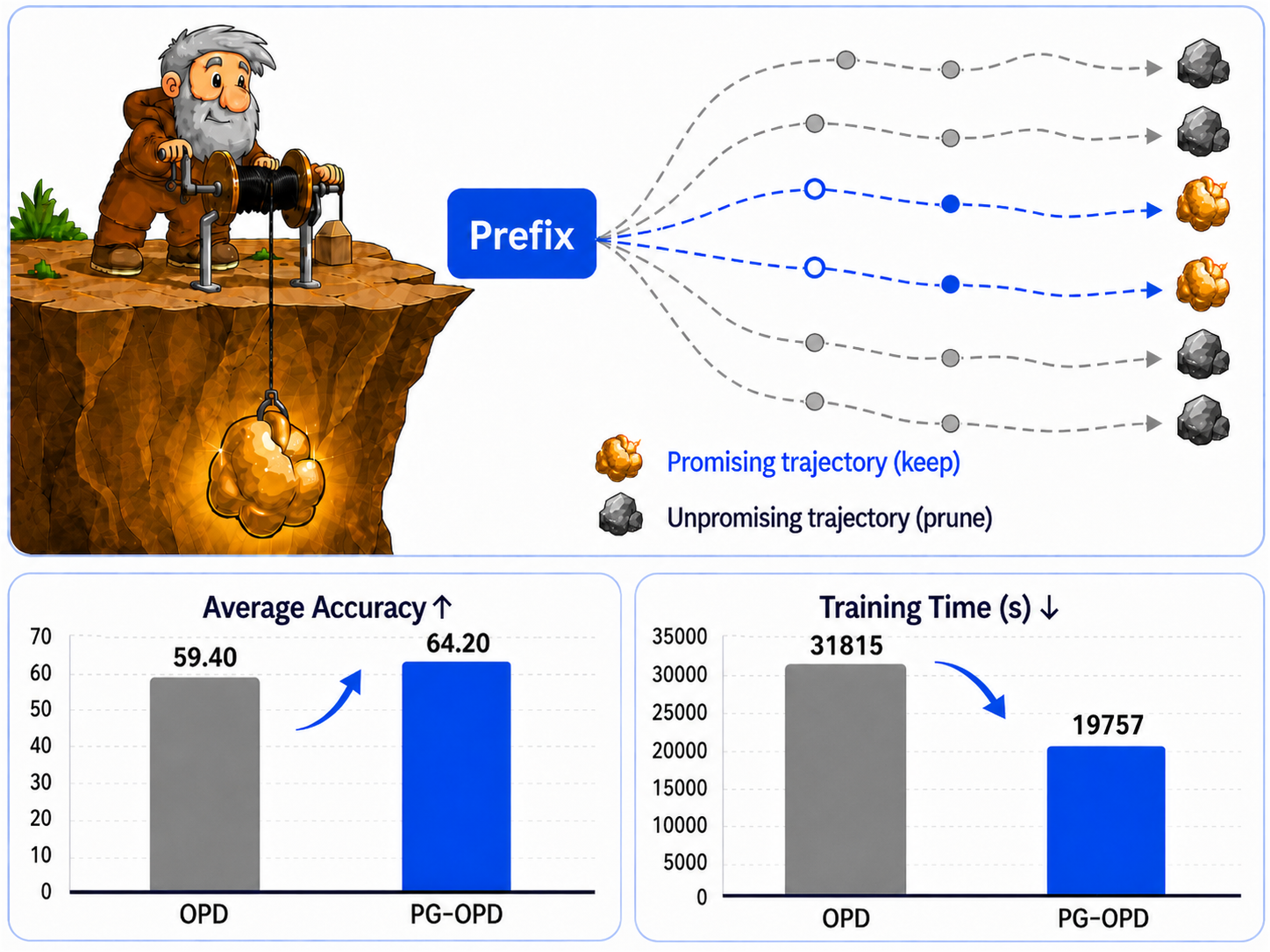}
\caption{Overview of PG-OPD. Prefix-guided selection identifies high-value trajectories from early prefixes and allocates long-rollout budget to them, improving both avg@16 accuracy and training efficiency.}
\label{fig:first}
\end{figure}

Despite its effectiveness, OPD incurs substantial computational cost. A typical OPD update samples multiple candidate trajectories for each prompt and decodes every candidate to the same long rollout budget. In long-horizon reasoning, individual trajectories can span thousands of tokens, making rollout generation a major component of training cost. Existing reasoning post-training pipelines often rely on large-scale sampling to support exploration and provide dense learning signals \citep{shao2024deepseekmath, deepseekai2025deepseekr1, he2025justrl, yu2025dapo}. OPD introduces additional overhead because the teacher must also be evaluated at student-visited token states to provide token-level supervision. Standard OPD therefore allocates comparable generation and supervision cost
to every sampled trajectory, implicitly assuming that all candidates warrant
the same long-rollout budget. In practice, however, sampled trajectories can
exhibit substantially different levels of teacher--student compatibility,
calling this uniform allocation into question.

Recent analyses suggest that OPD effectiveness depends not only on teacher strength, but also on the local compatibility between the teacher and student on student-visited states \citep{li2026rethinking, fu2026revisiting, jang2026stable, jin2026entropy}. A practical proxy for such compatibility is teacher--student top-$k$ overlap,
which measures the extent to which the two models share the same
high-probability next-token candidates. High-overlap trajectories are more likely to remain in regions where the teacher provides supervision compatible with the student's current policy. In contrast, low-overlap trajectories may enter regions where the teacher's supervision is less readily exploitable. Generating and training on low-overlap trajectories requires additional rollout generation,
increasing computational cost while providing limited distillation benefit. These observations suggest that teacher--student compatibility can help distinguish trajectories with different utility for distillation.

This motivates a different perspective on efficient OPD: rollout computation should account for differences in the distillation value of sampled trajectories. Although standard OPD decodes every candidate to the same maximum length, their early trajectories can exhibit substantially different levels of teacher--student compatibility. Candidates with high local overlap are more likely to remain in regions where teacher supervision is compatible with the student's current policy, whereas continuing low-overlap candidates may cause their trajectories to
drift further from the teacher as generation proceeds, increasing computational
cost while providing limited distillation benefit. A fixed-length prefix therefore offers a natural decision point at which teacher--student overlap can be used to estimate whether a trajectory is worth continuing. This suggests a simple rollout-allocation strategy: use short probe prefixes to identify promising candidates, continue only those with higher prefix
compatibility, and apply OPD training exclusively to the selected trajectories.

We propose Prefix-Guided On-Policy Distillation (PG-OPD), a rollout-allocation
framework for improving both the efficiency and effectiveness of OPD. PG-OPD first generates a short probe prefix for each sampled candidate and measures teacher--student compatibility using top-$k$ overlap over the prefix. Candidates with higher prefix-level compatibility are then continued to the target rollout length, while the remaining candidates stop at the prefix boundary. Only selected candidates are used for OPD training,
whereas unselected probe prefixes are discarded before
optimization. PG-OPD preserves the standard per-token reverse-KL loss while
inducing a compatibility-conditioned training distribution through
prefix-dependent selection.

We evaluate PG-OPD across multiple teacher--student combinations on five mathematical reasoning benchmarks from AMC, AIME, and HMMT. Across these settings, PG-OPD improves the accuracy--efficiency trade-off of standard OPD, substantially reducing training time while preserving or improving average benchmark performance at suitable rollout budgets. These results indicate that uniform long-rollout allocation is not always necessary for effective OPD. Instead, compatibility measured from short probe prefixes provides a practical and lightweight signal for directing rollout computation toward candidates that are more likely to provide useful distillation supervision.

Our contributions are summarized as follows:
\begin{itemize}
\item We formulate efficient OPD as a candidate-level rollout-allocation
problem and introduce PG-OPD, which uses short-prefix teacher--student
top-$k$ overlap to allocate a shared long-rollout budget across candidates
while guaranteeing per-prompt coverage.
\item We evaluate PG-OPD across multiple teacher--student pairs and five
mathematical reasoning benchmarks, demonstrating consistent improvements in
both reasoning performance and training efficiency over standard
OPD.
\item We analyze how selecting candidates with different prefix-overlap scores
affects final reasoning performance, showing that early teacher--student
compatibility is an effective signal for rollout allocation. 
\end{itemize}

\section{Related Work}

\paragraph{On-Policy Distillation}

Knowledge distillation transfers knowledge from a teacher model to a student model through distribution matching or teacher-generated trajectories \citep{hinton2015distilling, kim2016sequence}. Recent work extends this paradigm to on-policy distillation (OPD), where supervision is computed on trajectories sampled from the current student policy rather than on fixed teacher-generated data \citep{gu2024minillm, agarwal2024gkd}. By evaluating teacher distributions on student-visited states, OPD reduces train--inference mismatch and provides dense token-level learning signals throughout generation. This formulation has become an important component of modern post-training pipelines for reasoning models \citep{ yang2025qwen3, deepseekai2025deepseekr1, yu2025dapo}. OPD has also been extended to self-distillation and online learning settings, where models improve through their own generated trajectories and feedback signals \citep{hubotter2026rlsd, shenfeld2026selfdistillation, zhao2026selfdistilled}. Our work explores prefix-guided rollout allocation to improve both the
efficiency and effectiveness of OPD.

\paragraph{Teacher--Student Compatibility in OPD.}
Recent studies have shown that OPD effectiveness depends strongly on the
quality and compatibility of teacher supervision on student-visited states.
Teacher--student reasoning-pattern compatibility and top-$k$ token overlap
have been identified as key indicators of successful distillation
\citep{li2026rethinking}, while teacher guidance may become unreliable when
student-generated prefixes drift from the teacher's support
\citep{fu2026revisiting}.
Other work studies instability caused by large teacher--student distribution
gaps or teacher uncertainty
\citep{jang2026stable,jin2026entropy}.
Complementary approaches examine non-uniform OPD signals through token
importance, trajectory correctness, token teachability, and position-dependent
teacher reliability
\citep{xu2026tip,zheng2026scope,wang2026teachability,
liu2026positionweighted}.
Collectively, these studies show that teacher supervision is not equally useful
across student-visited states. However, these methods mainly analyze or adapt supervision at the token or
objective level, while still requiring full trajectory generation and teacher
evaluation. As a result, they do not directly reduce the computational cost of
long-horizon OPD.

\paragraph{Improving On-Policy Distillation.}
Recent work on OPD improves either training effectiveness or computational
efficiency. Objective-level methods reformulate the distillation target or
model teacher uncertainty for more stable optimization
\citep{jang2026stable,jin2026entropy,ko2026relaxed}.
Efficiency-oriented methods eliminate online teacher inference or shorten
student rollouts
\citep{wu2026lightning,zhang-etal-2026-fast,zhang2026fullrollouts,ziheng2026moreearlystoppingrollout,yang2026pruneopd}.
Lightning OPD precomputes teacher log-probabilities, while POPD, TOPD,
prefix distillation, and ESR control the generation or supervision horizon
of individual trajectories. PRUNE-OPD similarly adapts supervision and
termination online within each trajectory. In contrast, PG-OPD uses prefix compatibility to allocate a finite
continuation budget across multiple candidates sampled from the same prompt.
It selects a candidate set rather than per-trajectory stopping horizons,
while enforcing per-prompt coverage before global allocation. This reframes
efficient OPD from controlling individual rollout horizons to allocating
long-rollout computation across candidates.

\section{Method}

\begin{figure*}[t]
\centering
\includegraphics[width=\textwidth]{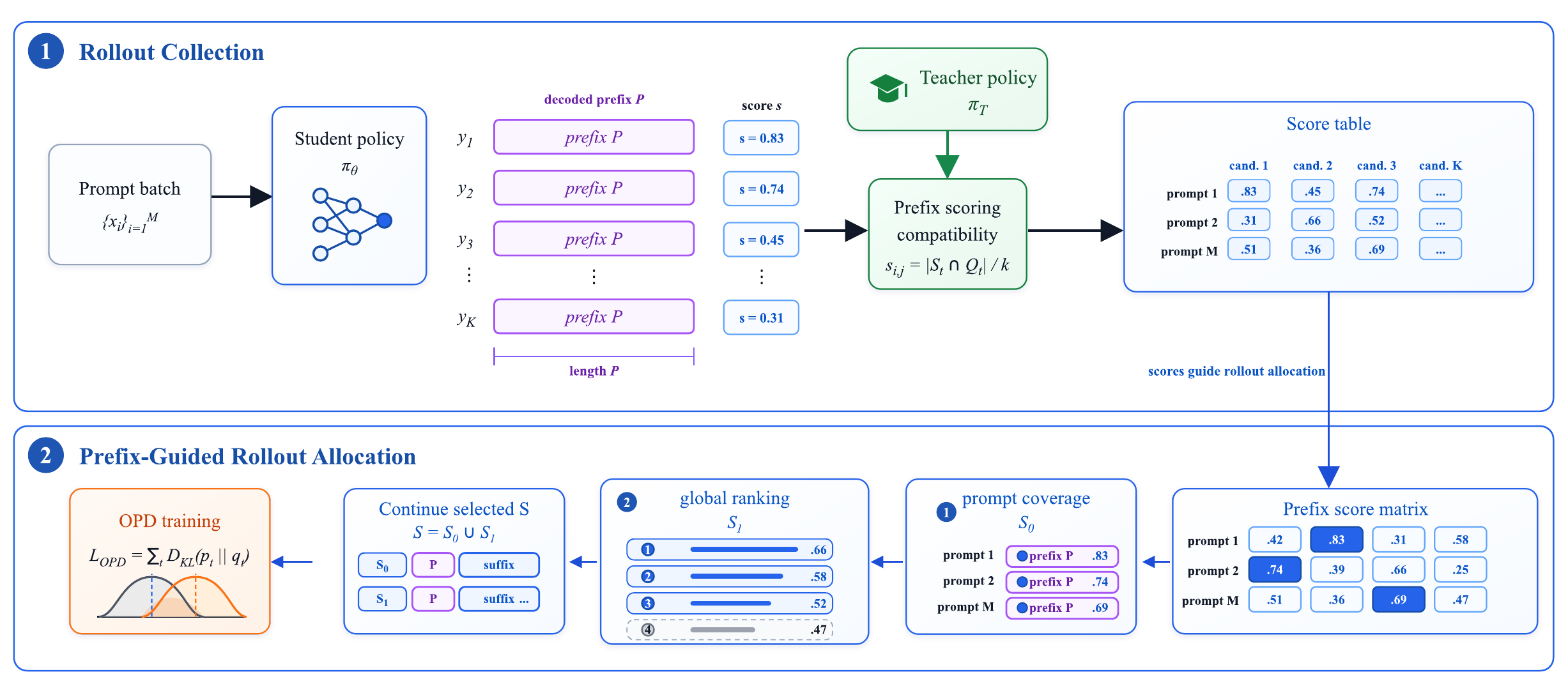}
\caption{Overview of PG-OPD. First, the student generates fixed-length prefixes
scored by teacher--student top-$k$ overlap. PG-OPD then forms \(S_0\) from
the highest-scoring candidate per prompt and \(S_1\) by globally ranking
the rest. Finally, the selected set \(S=S_0\cup S_1\) is continued to the target
rollout length and optimized using the standard per-token reverse-KL loss.}
\label{fig:pg_opd_overview}
\end{figure*}

\subsection{On-Policy Distillation}

Let $\pi_{\theta}$ and $\pi_{\mathrm{T}}$ denote the student and teacher
policies, respectively. Given a prompt $x$, standard on-policy distillation
samples a response from the current student policy:
\begin{equation}
y=(y_1,\ldots,y_T)
\sim
\pi_{\theta}(\cdot\mid x).
\end{equation}

At response position $t$, the student and teacher are evaluated on the same
student-generated context
$c_t=(x,y_{<t})$:
\begin{equation}
p_t(v)
=
\pi_{\theta}(v\mid c_t),
\qquad
q_t(v)
=
\pi_{\mathrm{T}}(v\mid c_t).
\end{equation}

The ideal full-vocabulary OPD objective is the reverse KL
divergence:
\begin{equation}
\ell^{\mathrm{OPD}}_t
=
D_{\mathrm{KL}}(p_t\Vert q_t)
=
\sum_{v\in\mathcal{V}}
p_t(v)
\log
\frac{p_t(v)}{q_t(v)}.
\label{eq:token_opd_loss}
\end{equation}

For a batch of $M$ prompts with $K$ sampled candidates per prompt, standard
OPD decodes every candidate to a common rollout length $L$ and minimizes
\begin{equation}
\mathcal{L}_{\mathrm{OPD}}
=
\frac{1}{Z_{\mathrm{OPD}}}
\sum_{i=1}^{M}
\sum_{j=1}^{K}
\sum_{t=1}^{L}
m_{i,j,t}
\ell^{\mathrm{OPD}}_{i,j,t},
\label{eq:opd_objective}
\end{equation}
where $m_{i,j,t}$ is the valid-response-token mask and
$Z_{\mathrm{OPD}}=\sum_{i,j,t}m_{i,j,t}$.

Standard OPD therefore assigns the same rollout length to every sampled
candidate. PG-OPD preserves the same per-token reverse-KL loss, while
prefix-dependent selection changes both rollout allocation and the trajectory
distribution contributing to optimization.

\subsection{Prefix Compatibility Scoring}

For each prompt $x_i$, PG-OPD first samples $K$ candidates from the student
and decodes each candidate to a probe-prefix length $P$:
\begin{equation}
y_{i,j,1:P}
\sim
\pi_{\theta}(\cdot\mid x_i),
\qquad
j=1,\ldots,K.
\end{equation}

At each valid prefix position $t\le P$, we evaluate the student and teacher
next-token distributions on the same student-generated context:
\begin{equation}
\begin{aligned}
p_{i,j,t}
&=
\pi_{\theta}
(\cdot\mid x_i,y_{i,j,<t}),\\
q_{i,j,t}
&=
\pi_{\mathrm{T}}
(\cdot\mid x_i,y_{i,j,<t}).
\end{aligned}
\end{equation}

Assuming a shared tokenizer and vocabulary between the teacher and student, we measure their local compatibility using top-k overlap:
\begin{equation}
o_{i,j,t}
=
\frac{
\left|
\operatorname{TopK}(p_{i,j,t},k)
\cap
\operatorname{TopK}(q_{i,j,t},k)
\right|
}{k}.
\label{eq:token_overlap}
\end{equation}

The compatibility score of candidate $(i,j)$ is the mask-normalized average
over its probe prefix:
\begin{equation}
s_{i,j}
=
\frac{
\sum_{t=1}^{P}
m_{i,j,t}o_{i,j,t}
}{
\sum_{t=1}^{P}
m_{i,j,t}
}.
\label{eq:prefix_score}
\end{equation}

A higher score indicates that the teacher and student share more of their
high-probability next-token candidates along the probe prefix. Motivated by
prior evidence that local teacher--student compatibility is informative of
distillation effectiveness \citep{li2026rethinking}, we use $s_{i,j}$ as a
lightweight candidate-selection signal. The score leaves the per-token
reverse-KL loss unchanged, but determines which candidates receive long
continuations and contribute to optimization.

\subsection{Prefix-Guided Rollout Allocation}

Consider a batch of $M$ prompts, with $K$ candidates sampled for each prompt.
Every candidate is first decoded to a probe length $P$ and assigned a
compatibility score $s_{i,j}$, where $i$ indexes prompts and $j$ indexes
candidates. Let $B$ denote the number of candidates selected for long
continuation and OPD training. PG-OPD guarantees that each prompt contributes
at least one selected candidate.

The selection proceeds in two stages. First, PG-OPD retains the
highest-scoring candidate for each prompt:
\begin{equation}
j_i^{*}
=
\arg\max_{1\le j\le K}s_{i,j},
\qquad
\mathcal{S}_0
=
\{(i,j_i^{*})\}_{i=1}^{M}.
\label{eq:prompt_min_selection}
\end{equation}
Requiring at least one selected candidate per prompt prevents the global
ranking from concentrating all continuation slots on a small subset of prompts.

The remaining $B-M$ slots form $\mathcal{S}_1$, which contains the
highest-scoring candidates not already in $\mathcal{S}_0$:
\begin{equation}
\begin{gathered}
\mathcal{S}_1=
\operatorname{Top}_{B-M}
\bigl(\{(i,j):(i,j)\notin\mathcal{S}_0\};s_{i,j}\bigr),\\
\mathcal{S}=\mathcal{S}_0\cup\mathcal{S}_1.
\end{gathered}
\label{eq:global_selection}
\end{equation}
where $\operatorname{Top}_{r}(\mathcal{A};s)$ selects the $r$
highest-scoring candidates from $\mathcal{A}$. The final selected
set $\mathcal{S}$ contains $B$ candidates.

Candidates in $\mathcal{S}$ are continued from the probe length $P$ to a total
response length $L$. Their complete trajectories
are retained for OPD training. Candidates outside $\mathcal{S}$ stop after the
probe stage and are discarded before optimization. The resulting training loss is
\begin{equation}
\mathcal{L}_{\mathrm{PG\mbox{-}OPD}}
=
\frac{
\displaystyle
\sum_{(i,j)\in\mathcal{S}}
\sum_{t=1}^{L}
m_{i,j,t}\,
\ell^{\mathrm{OPD}}_{i,j,t}
}{
\displaystyle
\sum_{(i,j)\in\mathcal{S}}
\sum_{t=1}^{L}
m_{i,j,t}
}.
\label{eq:pg_opd_objective}
\end{equation}
PG-OPD preserves the standard per-token reverse-KL loss.
However, prefix-dependent selection changes the trajectory
distribution contributing to optimization, thereby inducing a
compatibility-conditioned training distribution.

\begin{algorithm}[t]
\small
\caption{Prefix-Guided On-Policy Distillation}
\label{alg:pg_opd}
\begin{algorithmic}[1]
\REQUIRE Prompts $\{x_i\}_{i=1}^{M}$, student $\pi_\theta$,
teacher $\pi_{\mathrm{T}}$, candidates per prompt $K$,
probe length $P$, rollout length $L$, selected-candidate budget $B$, where $M\leq B\leq MK$
\STATE Generate a $P$-token probe prefix for each of the $MK$ candidates
\STATE Compute the compatibility score $s_{i,j}$ for each candidate using
Eq.~\eqref{eq:prefix_score}
\STATE Select the highest-scoring candidate for each prompt:
\[
\mathcal{S}_0
=
\{(i,\arg\max_j s_{i,j})\}_{i=1}^{M}
\]
\STATE Select the remaining $B-M$ candidates by global score ranking:
\[
\mathcal{S}
=
\mathcal{S}_0
\cup
\operatorname{Top}_{B-M}
\left(
\{(i,j):(i,j)\notin\mathcal{S}_0\};
s_{i,j}
\right)
\]
\FOR{each selected candidate $(i,j)\in\mathcal{S}$}
    \STATE Continue the probe prefix from length $P$ to a total length $L$
\ENDFOR
\STATE Discard the probe prefixes of all candidates not in $\mathcal{S}$
\STATE Update $\pi_\theta$ using the standard per-token reverse-KL loss on the selected
trajectories, including their probe-prefix tokens
\RETURN Updated student policy $\pi_\theta$
\end{algorithmic}
\end{algorithm}

\begin{table*}[t]
\centering
\small
\renewcommand{\arraystretch}{1.25}
\begin{tabular*}{\textwidth}{@{\extracolsep{\fill}}lcrrrrrrcc@{}}
\hline
Method & Pruned & AIME24 & AIME25 & AMC23 & HMMT24 & HMMT25 & Avg. &
\begin{tabular}{c}
Training \\
Time (s)
\end{tabular} & Speedup \\
\hline

\multicolumn{10}{l}{
\textit{OpenMath-Nemotron-1.5B / JustRL-Nemotron-1.5B}
} \\

OPD
& 0\%
& 68.96 & 60.21 & 96.56 & 30.83 & 40.42 & 59.40
& 31815 & 1.00$\times$ \\
OPD (Truncate 5K)
& --
& 65.83 & 63.54 & 95.62 & 36.04 & 33.12 & 58.83
& 22624 & 1.41$\times$ \\
PRUNE-OPD
& --
& 66.88 & 62.50 & 95.00 & 34.38 & 40.16 & 59.78
& 19933 & 1.60$\times$ \\
PG-OPD
& 25\%
& 73.13 & 64.17 & \textbf{97.34} & 36.25 & 39.17 & 62.01
& 21905 & 1.45$\times$ \\
PG-OPD
& 50\%
& \textbf{76.25} & \textbf{67.08} & 95.78 & 36.88 & \textbf{45.00}
& \textbf{64.20} & 19757 & 1.61$\times$ \\
PG-OPD
& 75\%
& 69.17 & 57.08 & 96.72 & \textbf{39.17} & 40.21 & 60.47
& \textbf{17008} & \textbf{1.87$\times$} \\

\hline
\multicolumn{10}{l}{
\textit{DeepSeek-R1-Distill-Qwen-1.5B / Skywork-OR1-Math-7B}
} \\
OPD
& 0\%
& 30.42 & \textbf{31.87} & \textbf{80.31} & 16.46 & 19.79 & 35.77
& 33023 & 1.00$\times$ \\
OPD (Truncate 5K)
& --
& 36.67 & 27.29 & 76.56 & \textbf{21.25} & 15.62 & 35.48
& 28416 & 1.16$\times$ \\
PRUNE-OPD
& --
& 30.63 & 25.21 & 69.84 & 11.67 & 17.29 & 30.93
& \textbf{6732} & \textbf{4.91$\times$} \\
PG-OPD
& 25\%
& \textbf{44.79} & 27.71 & 76.72 & 17.71 & 18.33 & \textbf{37.05}
& 23346 & 1.41$\times$ \\
PG-OPD
& 50\%
& 35.62 & 31.67 & 75.31 & 16.67 & \textbf{25.42} & 36.94
& 20759 & 1.59$\times$ \\
PG-OPD
& 75\%
& 38.54 & 31.04 & 71.88 & 18.96 & 21.67 & 36.42
& 18135 & 1.82$\times$ \\

\hline
\multicolumn{10}{l}{
\textit{DeepSeek-R1-Distill-Qwen-1.5B /
DeepSeek-R1-Distill-Qwen-7B}
} \\
OPD
& 0\%
& 30.63 & 24.58 & 70.94 & 10.42 & 16.67 & 30.65
& 38896 & 1.00$\times$ \\
OPD (Truncate 5K)
& --
& 36.25 & 23.33 & 72.50 & 11.90 & 15.95 & 31.99
& 24791 & 1.57$\times$ \\
PRUNE-OPD
& --
& 28.96 & 23.75 & 68.91 & 12.71 & 11.29 & 29.12
& \textbf{12868} & \textbf{3.02$\times$} \\
PG-OPD
& 25\%
& 33.75 & \textbf{27.08} & \textbf{74.38} & \textbf{13.75}
& \textbf{22.29} & \textbf{34.25} & 23181 & 1.68$\times$ \\
PG-OPD
& 50\%
& \textbf{38.96} & 26.25 & 70.00 & 12.08 & 17.71 & 33.00
& 20516 & 1.90$\times$ \\
PG-OPD
& 75\%
& 26.46 & 24.38 & 73.12 & 11.04 & 18.12 & 30.63
& 18294 & 2.13$\times$ \\

\hline
\end{tabular*}
\caption{
Main results across three teacher--student settings. Each block is formatted
as student / teacher. Accuracy is reported as avg@16 on five mathematical
reasoning benchmarks. ``Pruned'' denotes the fraction of candidates discarded
after prefix scoring, and speedup is computed relative to OPD within the same
setting.
}
\label{tab:ds15b_results}
\end{table*}

\begin{table*}[t]
\centering
\small
\renewcommand{\arraystretch}{1.25}
\begin{tabular*}{\textwidth}{@{\extracolsep{\fill}}lcrrrrrrcc@{}}
\hline
Method & Pruned & AIME24 & AIME25 & AMC23 & HMMT24 & HMMT25 & Avg. &
\begin{tabular}{c}
Training \\
Time (s)
\end{tabular} & Speedup \\
\hline

\multicolumn{10}{l}{
\textit{DeepSeek-R1-Distill-Qwen-1.5B / JustRL-DeepSeek-1.5B}
} \\
OPD
& 0\%
& 47.92 & 35.42 & 85.47 & 18.75 & 22.29 & 41.97
& 38221 & 1.00$\times$ \\
OPD (Truncate 5K)
& --
& 47.71 & 35.83 & \textbf{85.63} & 22.50 & 18.33 & 42.00
& 26725 & 1.43$\times$ \\
PRUNE-OPD
& --
& 47.08 & 31.46 & 82.65 & 20.21 & 18.33 & 39.95
& \textbf{15838} & \textbf{2.41$\times$} \\
PG-OPD
& 25\%
& 49.17 & 35.00 & 83.28 & 22.29 & 20.21 & 41.99
& 21883 & 1.75$\times$ \\

PG-OPD
& 50\%
& 44.17 & \textbf{36.25} & 85.47 & \textbf{24.17} & 20.42 & 42.09
& 20926 & 1.83$\times$ \\
PG-OPD
& 75\%
& \textbf{52.08} & 33.96 & 82.81 & 21.67 & \textbf{23.54}
& \textbf{42.81} & 17265 & 2.21$\times$ \\

\hline
\multicolumn{10}{l}{
\textit{Qwen3-4B-Base / Qwen3-4B}
} \\
OPD
& 0\%
& 16.46 & 16.04 & 48.91 & 5.83 & 4.17 & 18.28
& 62476 & 1.00$\times$ \\
OPD (Truncate 5K)
& --
& 14.58 & 13.75 & 48.12 & 7.50 & \textbf{8.75} & 18.54
& 39478 & 1.58$\times$ \\
PRUNE-OPD
& --
& 14.58 & 13.54 & 46.56 & 10.00 & 3.96 & 17.73
& \textbf{15143} & \textbf{4.13$\times$} \\
PG-OPD
& 25\%
& \textbf{17.08} & \textbf{16.88} & \textbf{57.32}
& \textbf{11.25} & 7.50 & \textbf{22.01}
& 36246 & 1.72$\times$ \\
PG-OPD
& 50\%
& 13.12 & 15.00 & 53.28 & 7.08 & 7.71 & 19.24
& 30609 & 2.04$\times$ \\
PG-OPD
& 75\%
& 16.25 & 12.08 & 54.84 & 5.00 & 8.33 & 19.30
& 25351 & 2.46$\times$ \\

\hline
\end{tabular*}
\caption{
Main results for the other two teacher--student settings.
Each block is formatted as student / teacher.
The evaluation protocol and column definitions follow
Table~\ref{tab:ds15b_results}.
}
\label{tab:additional_student_results}
\end{table*}

\section{Experiments}

\subsection{Setup}

We evaluate PG-OPD on five teacher--student pairs across diverse model
configurations:
DeepSeek-R1-Distill-Qwen-1.5B / JustRL-DeepSeek-1.5B
\citep{deepseekai2025deepseekr1,he2025justrl},
OpenMath-Nemotron-1.5B / JustRL-Nemotron-1.5B
\citep{moshkov2025openmathreasoning,he2025justrl},
DeepSeek-R1-Distill-Qwen-1.5B /
DeepSeek-R1-Distill-Qwen-7B
\citep{deepseekai2025deepseekr1},
DeepSeek-R1-Distill-Qwen-1.5B /
Skywork-OR1-Math-7B
\citep{deepseekai2025deepseekr1,he2025skywork}, and
Qwen3-4B-Base / Qwen3-4B
\citep{yang2025qwen3},
where each pair is formatted as student / teacher.
The evaluation covers five mathematical reasoning benchmarks:
AIME24 and AIME25 \citep{maa2026aime},
AMC23 \citep{maa2026amc}, and
HMMT24 and HMMT25 \citep{hmmt2026archive}.
We report avg@16 accuracy by averaging correctness over 16 sampled solutions
for each problem.

All methods are trained on the same DAPO-Math-17K dataset
\citep{yu2025dapo} with identical optimization steps and sampling settings. All OPD-based methods use the same student-top-16 approximation to the reverse-KL objective.
Unless otherwise noted, all ablation studies use the
DeepSeek-R1-Distill-Qwen-1.5B / JustRL-DeepSeek-1.5B setting. Both the rollout
sampling temperature and teacher temperature are set to $1.0$. The maximum
prompt length is 1,024 tokens, and the maximum training response length is
7,168 tokens for all experiments unless a method explicitly modifies the
effective rollout budget. All training runs are conducted on a single node
with eight NVIDIA H100 GPUs using the same hardware configuration. Complete
implementation details and hyperparameter settings are provided in the
supplementary material.

Unless otherwise specified, PG-OPD generates a $P=128$-token
probe prefix for each sampled candidate and scores it using
teacher--student top-16 overlap. Candidates with higher prefix
scores are selected for continuation to a total length of
$L=5000$ tokens and retained for OPD training, while the remaining
candidates are discarded after the probe stage. We evaluate pruning
ratios of 25\%, 50\%, and 75\%, defined as the fraction of candidates
discarded after prefix scoring. To maintain prompt-level coverage,
PG-OPD retains at least one candidate for each prompt.

  For evaluation, we sample 16 solutions per problem with temperature $0.7$,
  top-$p=0.95$, and a maximum generation length of 31744 tokens. We report mean
  rollout accuracy over the 16 sampled solutions. Training time is measured as
  wall-clock time in seconds under the same training setup.

\subsection{Compared Methods}

\noindent\textbf{OPD.}
The baseline follows standard on-policy distillation. It samples trajectories
from the current student policy, decodes every sampled trajectory to the full
response budget, and trains on all generated tokens.

\noindent\textbf{Truncated OPD.}
This baseline follows standard OPD but sets the maximum response length to
5,000 tokens.

\noindent\textbf{PRUNE-OPD.}
PRUNE-OPD~\citep{yang2026pruneopd} continuously monitors teacher--student
top-$k$ overlap during generation, progressively downweights supervision after
repeated low-overlap events, and uses the resulting reliable-length statistics
to dynamically adjust the rollout budget.

\noindent\textbf{PG-OPD.}
Prefix-Guided OPD first generates a short prefix for each sampled candidate,
uses teacher--student top-$k$ overlap on the prefix to estimate trajectory
value, and then allocates long continuations only to selected candidates.

\begin{figure}[t]
  \centering
  \includegraphics[width=\columnwidth]{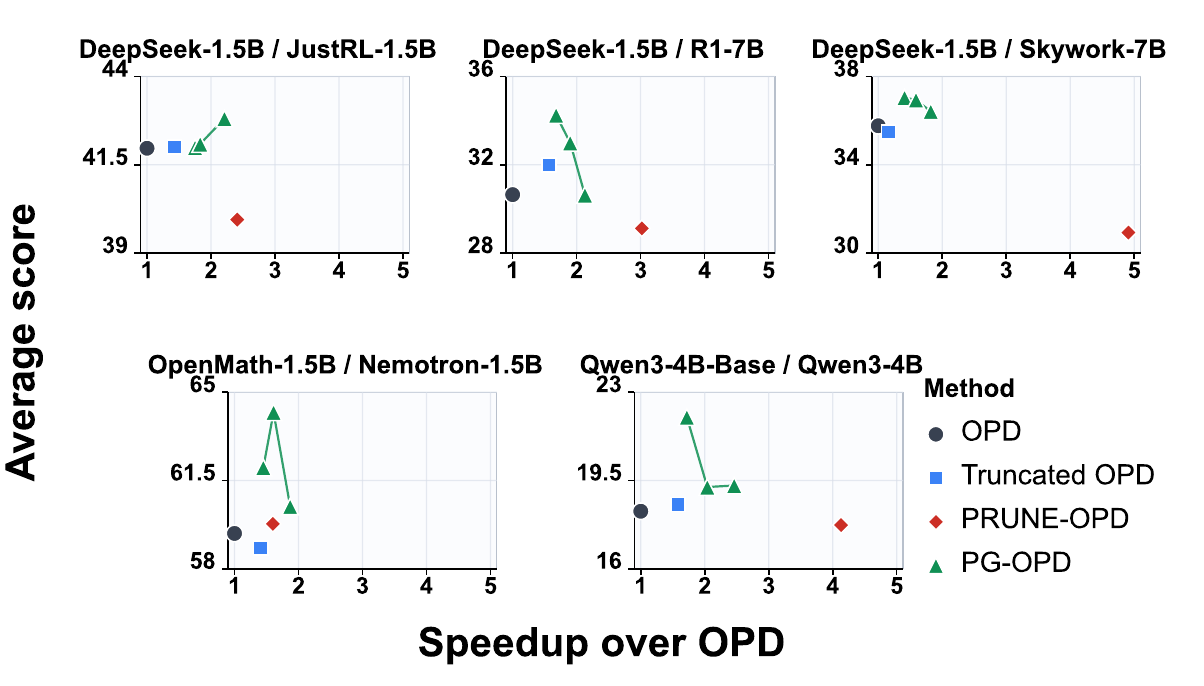}
\caption{Accuracy--efficiency trade-off across teacher--student pairs. Each point reports the average score over five benchmarks and the wall-clock speedup relative to standard OPD. PG-OPD yields stronger Pareto operating points by improving or preserving accuracy while reducing training time, whereas PRUNE-OPD often achieves larger speedups at the cost of lower accuracy.}
  \label{fig:accuracy_efficiency_pareto}
\end{figure}

\subsection{Main Results}

Tables~\ref{tab:ds15b_results} and
\ref{tab:additional_student_results} report results across five
teacher--student settings. Overall, PG-OPD shows that pruning low-overlap candidates can
reduce training cost while preserving or improving reasoning performance. A representative result appears in the
\textit{OpenMath-Nemotron-1.5B / JustRL-Nemotron-1.5B} setting, where PG-OPD with
50\% pruning improves the average score from 59.40 to 64.20, a gain of
4.80 points, while reducing training time from 31,815 to 19,757 seconds and
achieving a 1.61$\times$ speedup. In comparison, PRUNE-OPD often obtains
larger speedups but can incur accuracy degradation. This is evident in the
\textit{DeepSeek-R1-Distill-Qwen-1.5B / Skywork-OR1-Math-7B} setting, where
PRUNE-OPD achieves a 4.91$\times$ speedup but reduces the average score from
35.77 to 30.93.

These results suggest that selectively pruning low-overlap rollouts can
improve training efficiency while generally preserving or enhancing
reasoning performance. The observed improvements are consistent with two
coupled effects: PG-OPD avoids long continuations for low-overlap candidates
and shifts the training distribution toward trajectories with stronger
prefix-level compatibility.

\subsection{Uniform Rollout-Length Ablation}

\begin{figure}[t]
\centering
\includegraphics[width=\columnwidth]{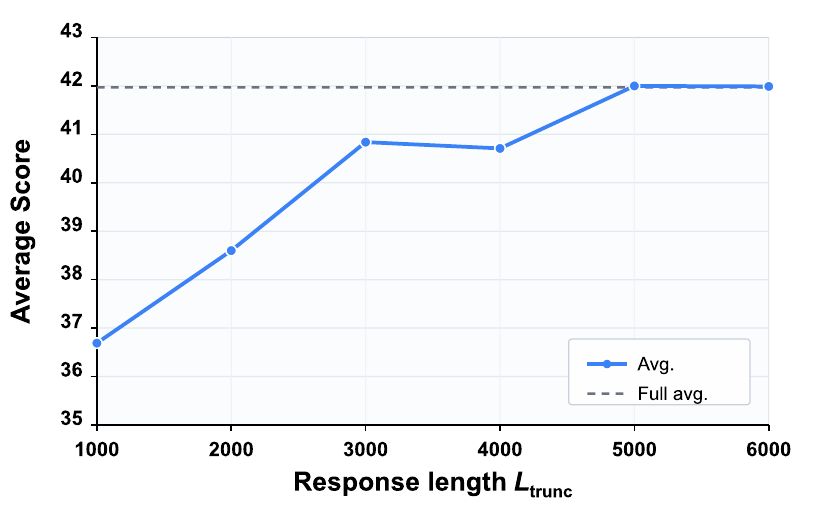}
\caption{Uniform rollout-length ablation. All trajectories are trained under
the same maximum response length $L_{\mathrm{trunc}}$ to evaluate the effect
of uniform truncation on performance.}
\label{fig:prefix_length_ablation}
\end{figure}

Figure~\ref{fig:prefix_length_ablation} studies uniformly truncated OPD by
restricting every sampled trajectory to a maximum response length
$L_{\mathrm{trunc}}$. The average score increases from 36.69 at
$L_{\mathrm{trunc}}=1000$ to 42.00 at $L_{\mathrm{trunc}}=5000$ and remains
nearly unchanged at 41.99 for $L_{\mathrm{trunc}}=6000$, both comparable to
the 41.97 achieved by full-length OPD. These results show that moderately
shortened rollouts can preserve the overall performance of full-length OPD
while reducing generation cost. However, the degradation under more
aggressive length limits indicates that uniform truncation may remove useful
long-horizon supervision required for difficult reasoning steps.

\subsection{Selection Signal Ablation}

\begin{table}[t]
\centering
\scriptsize
\setlength{\tabcolsep}{2.2pt}
\renewcommand{\arraystretch}{1.15}
\begin{tabular*}{\columnwidth}{@{\extracolsep{\fill}}lcrrrrrr@{}}
\hline
Method & Pruned & AIME24 & AIME25 & AMC23 & HMMT24 & HMMT25 & Avg. \\
\hline
OPD
& 0\%
& 68.96 & 60.21 & 96.56 & 30.83 & 40.42 & 59.40 \\
\hline

Random
& 25\%
& 68.54 & 59.79 & 96.25 & 32.92 & 37.08 & 58.92 \\
Loss-Based
& 25\%
& 67.92 & 60.42 & 95.47 & 32.29 & 36.67 & 58.55 \\
PG-OPD
& 25\%
& \textbf{73.13} & \textbf{64.17} & \textbf{97.34}
& \textbf{36.25} & \textbf{39.17} & \textbf{62.01} \\
\hline

Random
& 50\%
& 67.29 & 59.17 & \textbf{95.94} & 32.50 & 36.88 & 58.35 \\
Loss-Based
& 50\%
& 67.71 & 59.58 & 95.31 & 32.08 & 37.29 & 58.40 \\
PG-OPD
& 50\%
& \textbf{76.25} & \textbf{67.08} & 95.78
& \textbf{36.88} & \textbf{45.00} & \textbf{64.20} \\
\hline

Random
& 75\%
& 66.67 & 58.33 & 95.16 & 31.46 & 36.04 & 57.53 \\
Loss-Based
& 75\%
& 66.25 & \textbf{58.75} & 94.69 & 31.88 & 36.46 & 57.60 \\
PG-OPD
& 75\%
& \textbf{69.17} & 57.08 & \textbf{96.72}
& \textbf{39.17} & \textbf{40.21} & \textbf{60.47} \\
\hline
\end{tabular*}
\caption{Comparison of random, loss-based, and overlap-based selection under
pruning ratios in the OpenMath-Nemotron-1.5B /
JustRL-Nemotron-1.5B setting.}
\label{tab:selection_signal_ablation}
\end{table}

Table~\ref{tab:selection_signal_ablation} compares random, loss-based, and
overlap-based candidate selection at matched pruning ratios. Random selection samples candidates uniformly without using a prefix-derived
ranking signal. Loss-Based retains candidates with the highest
mean reverse-KL loss over the probe prefix, while PG-OPD retains those with
the highest teacher--student top-$k$ overlap. PG-OPD achieves the highest
average score at all pruning ratios, with the largest margin at 50\%,
suggesting that prefix-level overlap is more effective than prefix loss for
allocating long-continuation budget.

\subsection{Prompt-Level Coverage Ablation}

Table~\ref{tab:prompt_min1_ablation} compares three candidate-allocation
strategies under 50\% pruning. Global Ranking selects candidates across the
entire batch, Intra-Prompt Selection independently retains the top two
candidates within each prompt, and Per-Prompt Coverage first retains the
highest-scoring candidate for every prompt before allocating the remaining
slots globally. Per-Prompt Coverage achieves the best average score of 42.09,
compared with 40.67 for Global Ranking and 41.22 for Intra-Prompt Selection.
These results suggest that combining prompt-level coverage with global
allocation provides a better balance between preserving prompt diversity and
selecting high-scoring candidates.

\begin{table}[t]
\centering
\scriptsize
\setlength{\tabcolsep}{2.2pt}
\renewcommand{\arraystretch}{1.15}
\begin{tabular*}{\columnwidth}{@{\extracolsep{\fill}}lrrrrrr@{}}
\hline
Method & AIME24 & AIME25 & AMC23 & HMMT24 & HMMT25 & Avg. \\
\hline

Global Ranking
& 43.96
& \textbf{36.67}
& 82.50
& 18.75
& \textbf{21.46}
& 40.67 \\

Per-Prompt Coverage
& 44.17
& 36.25
& \textbf{85.47}
& \textbf{24.17}
& 20.42
& \textbf{42.09} \\

Intra-Prompt Selection
& \textbf{46.25}
& 35.42
& 84.22
& 20.83
& 19.38
& 41.22 \\

\hline
\end{tabular*}
\caption{Performance comparison of Global Ranking, Per-Prompt Coverage, and
Intra-Prompt Selection under 50\% pruning across five mathematical reasoning
benchmarks.}
\label{tab:prompt_min1_ablation}
\end{table}

\subsection{Effect of Prefix-Based Ranking}

\begin{figure}[t]
\centering
\includegraphics[width=\columnwidth]{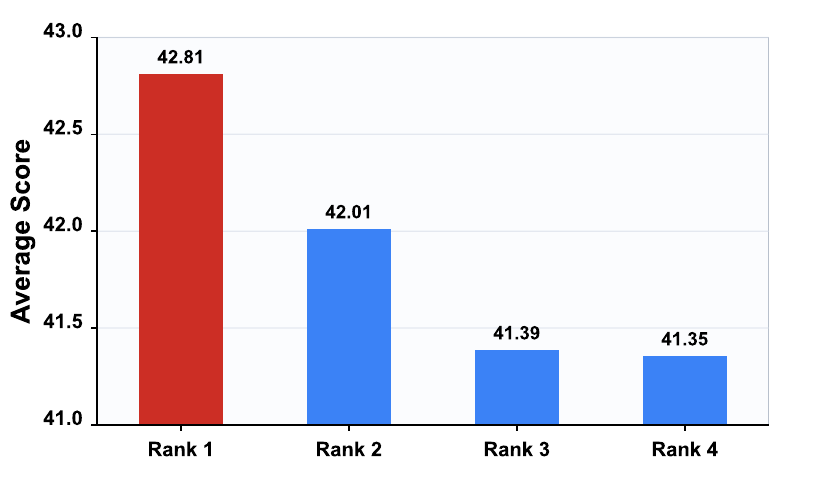}
\caption{Effect of prefix-based ranking under 75\% pruning. Each bar reports the average benchmark accuracy when long-continuation budget is allocated only to trajectories at a fixed prefix-overlap rank. }
\label{fig:rank_average_bar}
\end{figure}

To assess whether prefix-overlap scores provide a meaningful candidate
ordering, we compare rank-specific selection under 75\% pruning by retaining,
for each prompt, the candidate with a specified prefix-overlap rank for
training. Figure~\ref{fig:rank_average_bar} reports the average benchmark
accuracy obtained at each rank. Rank-1 candidates achieve the highest average
score, while candidates at ranks 2--4 perform worse. This result indicates
that overlap measured from short prefixes is informative for identifying
candidates that lead to more effective OPD training. Together with the random-
and loss-based selection baselines, these results indicate that prefix overlap
provides a more informative criterion for retaining effective training
trajectories.

\subsection{Overlap Scoring Window Ablation}

Table~\ref{tab:probe_token_ablation} studies the effect of probe length on
prefix-overlap scoring. A 32-token probe achieves an average score of 38.81,
suggesting that very short prefixes may not provide sufficient information for
reliable candidate selection. Increasing the probe length to 64 tokens improves
the average score to 42.05, while 128 tokens yields the best overall performance
of 42.81. The comparable result at 256 tokens suggests that a 128-token prefix
already provides sufficient information for candidate selection. 

\begin{table}[t]
\centering
\scriptsize
\setlength{\tabcolsep}{2.5pt}
\renewcommand{\arraystretch}{1.15}
\begin{tabular*}{\columnwidth}{@{\extracolsep{\fill}}lrrrrrr@{}}
\hline
Probe Tokens & AIME24 & AIME25 & AMC23 & HMMT24 & HMMT25 & Avg. \\
\hline
32  & 44.58 & 31.46 & 82.19 & 18.96 & 16.88 & 38.81 \\
64  & 44.38 & \textbf{35.42} & \textbf{85.47} &
      \textbf{22.71} & 22.29 & 42.05 \\
128 & \textbf{52.08} & 33.96 & 82.81 & 21.67 &
      \textbf{23.54} & \textbf{42.81} \\
256 & 50.83 & 34.58 & 83.44 & 21.46 & 23.33 & 42.73 \\
\hline
\end{tabular*}
\caption{Ablation on the number of probe tokens used to compute the
prefix-level overlap score. Probe Tokens denotes the number of initial tokens
used for teacher--student top-$k$ overlap estimation before making pruning
decisions.}
\label{tab:probe_token_ablation}
\end{table}

\section{Conclusion}

Standard OPD assigns the same long rollout budget to every student-sampled
candidate, creating substantial overhead in long-horizon reasoning. We
introduced Prefix-Guided On-Policy Distillation (PG-OPD), which uses
teacher--student top-$k$ overlap from short probe prefixes to select candidates
for long continuation and training. By discarding low-overlap candidates after the probe stage,
PG-OPD reduces unnecessary long-horizon generation while
retaining the standard per-token reverse-KL loss on selected
trajectories. Prefix-dependent selection nevertheless induces
a compatibility-conditioned training distribution. Across five
teacher--student settings and five mathematical reasoning benchmarks, PG-OPD
achieves up to a 4.80-point average-score improvement and a 2.46$\times$
wall-clock speedup. Ablation studies further support prefix-level overlap as
an effective signal for candidate selection and rollout allocation. These
results demonstrate that selectively pruning low-overlap candidates can
improve both training efficiency and reasoning performance across
different model configurations.

\bibliography{aaai2027}

\clearpage

\appendix
\onecolumn

\newtcolorbox{questionboxwide}{
  enhanced,
  breakable,
  colback=QuestionBack,
  colframe=QuestionFrame,
  boxrule=0.55pt,
  arc=1mm,
  left=1.5mm,
  right=1.5mm,
  top=1.2mm,
  bottom=1.2mm,
  before skip=3pt,
  after skip=6pt,
  fontupper=\small
}

\newtcolorbox{fullopdwidebox}{
  enhanced,
  breakable,
  colback=FullBack,
  colframe=FullFrame,
  boxrule=0.65pt,
  arc=1mm,
  title={\opd{} Response \xmark},
  fonttitle=\bfseries\small,
  fontupper=\footnotesize,
  left=1.5mm,
  right=1.5mm,
  top=1.2mm,
  bottom=1.2mm,
  before upper={
    \setlength{\abovedisplayskip}{3pt}
    \setlength{\belowdisplayskip}{3pt}
    \setlength{\abovedisplayshortskip}{3pt}
    \setlength{\belowdisplayshortskip}{3pt}
  }
}

\newtcolorbox{pgopdwidebox}{
  enhanced,
  breakable,
  colback=PGBack,
  colframe=PGFrame,
  boxrule=0.65pt,
  arc=1mm,
  title={\pgopd{} Response \cmark},
  fonttitle=\bfseries\small,
  fontupper=\footnotesize,
  left=1.5mm,
  right=1.5mm,
  top=1.2mm,
  bottom=1.2mm,
  before upper={
    \setlength{\abovedisplayskip}{3pt}
    \setlength{\belowdisplayskip}{3pt}
    \setlength{\abovedisplayshortskip}{3pt}
    \setlength{\belowdisplayshortskip}{3pt}
  }
}
\newtcolorbox{promptbox}[1]{
  colback=PromptBlueLight,
  colframe=PromptBlue,
  colbacktitle=PromptBlue,
  coltitle=white,
  fonttitle=\bfseries,
  title=#1,
  boxrule=0.9pt,
  arc=3pt,
  outer arc=3pt,
  left=8pt,
  right=8pt,
  top=7pt,
  bottom=7pt,
  toptitle=4pt,
  bottomtitle=4pt
}
\section{Prompt Template}
\label{app:templates}

The same prompt template is used for evaluation on AIME24, AIME25, AMC23, HMMT24, and HMMT25, with \texttt{{Question}} replaced by the corresponding problem statement.

\begin{promptbox}{Prompt template}
\ttfamily
\{Question\} Please reason step by step, and put your final answer within
\textbackslash boxed\{\}.
\end{promptbox}

\section{Training and Implementation Details}
\label{app:training_details}

All methods use the same optimization and sampling setup unless otherwise
specified. Table~\ref{tab:training_details} summarizes the main training
hyperparameters, and Table~\ref{tab:eval_details} summarizes the evaluation
configuration.

\begin{table}[h]
\centering
\normalsize
\setlength{\tabcolsep}{8pt}
\renewcommand{\arraystretch}{1.5}
\begin{tabular}{ll}
\toprule
\textbf{Hyperparameter} & \textbf{Value} \\
\midrule
Training dataset & DAPO-Math-17K \\
Number of training steps & 279 \\
Training epochs & 1 \\
Mini-batch size & 64 \\
Responses per prompt & 4 \\
Maximum prompt length & 1024 \\
Standard OPD maximum response length & 7168 \\
Optimizer learning rate & $1\times 10^{-6}$ \\
Rollout temperature & 1.0 \\
Teacher temperature & 1.0 \\
Repetition penalty & 1.0 \\
Loss approximation & Student top-16 \\
Prefix-overlap top-$k$ & 16 \\
Loss aggregation & Token mean \\
KL regularization & Disabled \\
Hardware & 8 NVIDIA H100 GPUs \\
\bottomrule
\end{tabular}
\caption{Training hyperparameters used for OPD, PG-OPD, and PRUNE-OPD.}
\label{tab:training_details}
\end{table}

\begin{table}[H]
\centering
\normalsize
\setlength{\tabcolsep}{7pt}
\renewcommand{\arraystretch}{1.22}
\begin{tabular}{p{0.42\textwidth}p{0.50\textwidth}}
\toprule
\textbf{Hyperparameter} & \textbf{Value} \\
\midrule
Evaluation benchmarks & AIME24, AIME25, AMC23, HMMT24, HMMT25 \\
Solutions per problem & 16 \\
Evaluation temperature & 0.7 \\
Top-$p$ & 0.95 \\
Maximum generation length & 31744 \\
Metric & avg@16 accuracy \\
\bottomrule
\end{tabular}
\caption{Evaluation configuration used for all reported benchmark results.}
\label{tab:eval_details}
\end{table}

\section{Additional Results}

\subsection{Generalization to Additional Model Pairs}
\label{app:additional_pairs}

Table~\ref{tab:additional_results} reports additional results on Qwen
teacher--student combinations. On Qwen3-1.7B-Base / Qwen3-4B, PG-OPD improves
the average score from 10.85 to 13.67 under 50\% pruning while achieving a
1.44$\times$ speedup. We also include Qwen3-4B-Base / Qwen3-8B as an
additional scale-up comparison.

\begin{table}[H]
\centering
\small
\renewcommand{\arraystretch}{1.22}
\begin{tabular*}{\textwidth}{@{\extracolsep{\fill}}lcrrrrrrcc@{}}
\toprule
Method & Pruned & AIME24 & AIME25 & AMC23 & HMMT24 & HMMT25 & Avg. &
\begin{tabular}{c}
Training \\
Time (s)
\end{tabular} & Speedup \\
\midrule

\multicolumn{10}{l}{\textit{Qwen3-1.7B-Base / Qwen3-4B}} \\
OPD & 0\%  & 8.96 & 3.96 & 37.81 & 3.54 & 0.00 & 10.85 & 46441 & 1.00$\times$ \\
PG-OPD & 25\% & 11.46 & 6.25 & 40.62 & 4.17 & 0.00 & 12.50 & 38100 & 1.22$\times$ \\
PG-OPD & 50\% & \textbf{13.12} & \textbf{7.29} & \textbf{43.12} & \textbf{4.79} & 0.00 & \textbf{13.67} & 32282 & 1.44$\times$ \\
PG-OPD & 75\% & 7.29 & 4.38 & 34.69 & 4.17 & 0.00 & 10.10 & 20341 & 2.28$\times$ \\

\midrule
\multicolumn{10}{l}{\textit{Qwen3-4B-Base / Qwen3-8B}} \\
OPD & 0\%  & 16.67 & 14.58 & \textbf{53.91} & 3.75 & 1.25 & 18.03 & 66486 & 1.00$\times$ \\
PG-OPD & 25\% & 16.67 & 11.25 & \textbf{53.91} & \textbf{8.33} & 2.50 & 18.53 & 36922 & 1.80$\times$ \\
PG-OPD & 50\% & \textbf{18.33} & 12.92 & 51.25 & 4.38 & \textbf{4.79} & 18.33 & 31017 & 2.14$\times$ \\
PG-OPD & 75\% & 14.17 & \textbf{18.10} & 52.97 & 7.71 & 1.67 & \textbf{18.92} & 25906 & \textbf{2.57$\times$} \\
\bottomrule
\end{tabular*}
\caption{Additional results on Qwen teacher--student combinations. Each block
is formatted as student / teacher. Accuracy is reported as avg@16 on five
mathematical reasoning benchmarks. ``Pruned'' denotes the fraction of
candidates discarded after prefix scoring. Speedup is computed relative to
OPD under the same teacher--student setting.}
\label{tab:additional_results}
\end{table}

\subsection{Student Performance Before Distillation}
\label{app:initial_student}

Table~\ref{tab:initial_student_performance} reports the performance of each
student model before OPD training, providing a common reference point for
interpreting the improvements obtained after distillation across different
teacher--student settings.

\begin{table}[H]
\centering
\small
\setlength{\tabcolsep}{3.2pt}
\renewcommand{\arraystretch}{2.0}
\begin{tabular*}{\textwidth}{@{\extracolsep{\fill}}p{0.28\textwidth}rrrrrr@{}}
\toprule
Student Model & AIME24 & AIME25 & AMC23 & HMMT24 & HMMT25 & Avg. \\
\midrule
DeepSeek-R1-Distill-Qwen-1.5B & 29.17 & 25.21 & 69.06 & 13.12 & 12.92 & 29.90 \\
Qwen3-1.7B-Base & 3.12 & 0.00 & 14.53 & 0.00 & 0.00 & 3.53 \\
Qwen3-4B-Base & 8.75 & 9.79 & 27.50 & 3.33 & 1.67 & 10.21 \\
OpenMath-Nemotron-1.5B & 53.12 & 48.12 & 88.12 & 29.58 & 32.92 & 50.38 \\
\bottomrule
\end{tabular*}
\caption{Student avg@16 performance before OPD training.}
\label{tab:initial_student_performance}
\end{table}

\subsection{Uniform Rollout-Length Ablation}
\label{app:rollout_length}

Table~\ref{tab:app_rollout_length_ablation} provides the detailed results for
uniformly truncated OPD. All sampled trajectories are trained using at most
the first $L_{\mathrm{trunc}}$ response tokens. Performance improves as the
maximum rollout length increases and reaches its best average score at 5,000
tokens. Further increasing the rollout length to 6,000 tokens or using
full-length responses provides no additional performance gain.

\begin{table}[H]
\centering
\small
\setlength{\tabcolsep}{4.0pt}
\renewcommand{\arraystretch}{1.5}
\begin{tabular*}{\textwidth}{@{\extracolsep{\fill}}lrrrrrr@{}}
\toprule
Max Length & AIME24 & AIME25 & AMC23 & HMMT24 & HMMT25 & Avg. \\
\midrule
1000 & 37.92 & 34.17 & 77.19 & 17.92 & 16.25 & 36.69 \\
2000 & \textbf{48.75} & 27.56 & 81.07 & 18.54 & 17.08 & 38.60 \\
3000 & 48.54 & 33.75 & 81.09 & 18.13 & \textbf{22.71} & 40.84 \\
4000 & 47.50 & 33.13 & 84.38 & 21.04 & 17.50 & 40.71 \\
5000 & 47.71 & \textbf{35.83} & \textbf{85.63} & 22.50 & 18.33 & \textbf{42.00} \\
6000 & 46.88 & 34.17 & 83.91 & \textbf{22.71} & 22.29 & 41.99 \\
\midrule
Full & 47.92 & 35.42 & 85.47 & 18.75 & 22.29 & 41.97 \\
\bottomrule
\end{tabular*}
\caption{Detailed uniform rollout-length ablation. Full denotes standard OPD
with the original maximum response length.}
\label{tab:app_rollout_length_ablation}
\end{table}

\subsection{Prefix-Based Rank Analysis}
\label{app:rank_analysis}

Table~\ref{tab:app_rank_analysis} reports the detailed rank analysis. Rank 1
denotes the candidate with the highest overlap score measured from the probe
prefix. Selecting Rank-1 candidates yields the best average performance.

\begin{table}[H]
\centering
\small
\setlength{\tabcolsep}{4.0pt}
\renewcommand{\arraystretch}{1.5}
\begin{tabular*}{\textwidth}{@{\extracolsep{\fill}}lrrrrrr@{}}
\toprule
Rank & AIME24 & AIME25 & AMC23 & HMMT24 & HMMT25 & Avg. \\
\midrule
1 & \textbf{52.08} & 33.96 & 82.81 & 21.67 & \textbf{23.54} & \textbf{42.81} \\
2 & 46.25 & \textbf{38.33} & 84.84 & 21.46 & 19.17 & 42.01 \\
3 & 44.38 & 36.25 & 83.59 & \textbf{23.13} & 19.58 & 41.39 \\
4 & 44.17 & 36.46 & \textbf{85.31} & 18.54 & 22.29 & 41.35 \\
\bottomrule
\end{tabular*}
\caption{Detailed prefix-based rank analysis. Each row reports performance
when long continuations are allocated to candidates at a fixed overlap rank.}
\label{tab:app_rank_analysis}
\end{table}

\subsection{Per-Step Training-Time Breakdown}
\label{app:time_breakdown}

\begin{figure}[H]
\centering

\begin{minipage}[t]{0.47\textwidth}
\vspace{0pt}
\centering
\begin{adjustbox}{valign=t}
\includegraphics[width=\linewidth]{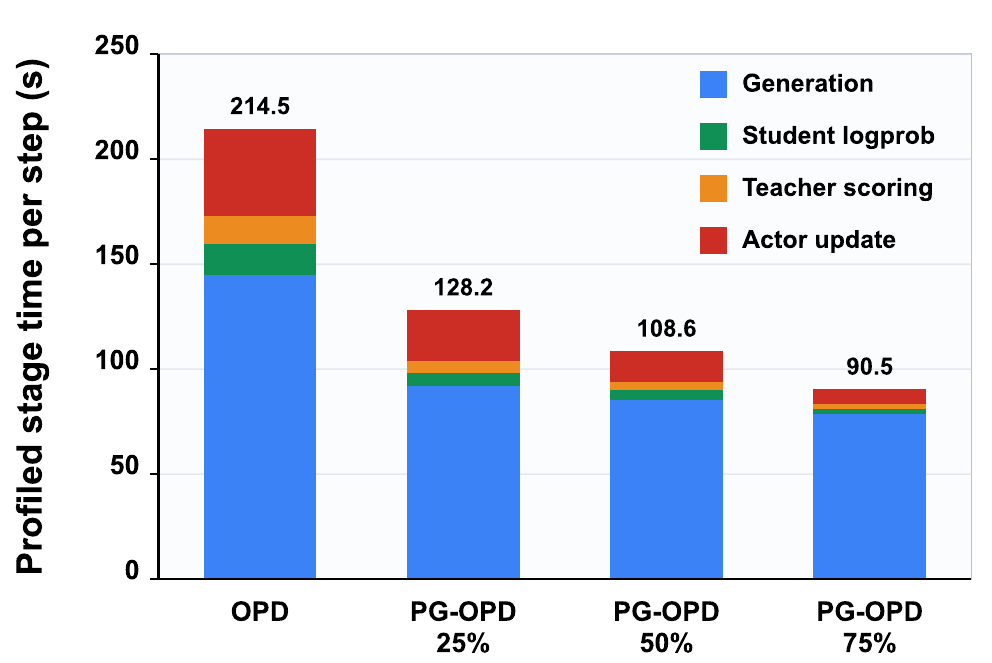}
\end{adjustbox}
\end{minipage}
\hfill
\begin{minipage}[t]{0.49\textwidth}
\vspace{8mm} 
\centering
\small
\setlength{\tabcolsep}{1.5pt}
\renewcommand{\arraystretch}{1.25}

\begin{tabular}{lrrrr}
\toprule
Method
& \makecell[c]{Rollout\\Generation}
& \makecell[c]{Student\\Log-Probability}
& \makecell[c]{Teacher\\Scoring}
& \makecell[c]{Actor\\Update} \\
\midrule
OPD
& 144.65 & 14.79 & 13.63 & 41.43 \\
PG-OPD-25\%
& 91.79 & 6.24 & 6.08 & 24.05 \\
PG-OPD-50\%
& 85.12 & 4.85 & 4.14 & 14.47 \\
PG-OPD-75\%
& 78.47 & 2.78 & 2.16 & 7.08 \\
\bottomrule
\end{tabular}
\end{minipage}

\caption{Per-step time breakdown of four profiled training stages for
OPD and PG-OPD on Qwen3-4B-Base / Qwen3-4B. The left panel shows the
decomposition, and the right panel reports the corresponding times.}
\label{fig:pg_opd_time_breakdown}
\end{figure}

Figure~\ref{fig:pg_opd_time_breakdown} reports the average per-step
time spent in four major training stages for Qwen3-4B-Base / Qwen3-4B
over one training epoch on DAPO-Math-17K: rollout generation, student
log-probability computation, teacher scoring, and actor updates.
The summed time of these profiled stages decreases from 214.5 seconds
per step for standard OPD to 128.2, 108.6, and 90.5 seconds for PG-OPD
with 25\%, 50\%, and 75\% pruning, respectively. These measurements
exclude auxiliary overhead outside the four profiled stages and therefore
do not directly correspond to the end-to-end wall-clock training times
reported in the main paper. Rollout generation remains the dominant
computational cost, while retaining fewer trajectories also substantially
reduces the subsequent student, teacher, and actor-update computation.

\providecommand{\casecmark}{%
  \textcolor{green!45!black}{\ding{51}}%
}
\providecommand{\casexmark}{%
  \textcolor{red!70!black}{\ding{55}}%
}

\definecolor{CaseQuestionBack}{RGB}{246,247,249}
\definecolor{CaseQuestionFrame}{RGB}{110,118,129}
\definecolor{CaseFullBack}{RGB}{253,242,242}
\definecolor{CaseFullFrame}{RGB}{185,65,65}
\definecolor{CasePGBack}{RGB}{241,249,243}
\definecolor{CasePGFrame}{RGB}{58,135,82}

\newtcolorbox{casequestionbox}{
  enhanced,
  breakable,
  colback=CaseQuestionBack,
  colframe=CaseQuestionFrame,
  boxrule=0.55pt,
  arc=1mm,
  left=2mm,
  right=2mm,
  top=1.5mm,
  bottom=1.5mm,
  before skip=4pt,
  after skip=6pt,
  fontupper=\small
}

\newtcolorbox{casefullopdbox}{
  enhanced,
  breakable,
  colback=CaseFullBack,
  colframe=CaseFullFrame,
  colbacktitle=CaseFullFrame,
  coltitle=white,
  boxrule=0.65pt,
  arc=1mm,
  title={OPD Response \casexmark},
  fonttitle=\bfseries\small,
  fontupper=\footnotesize,
  left=2mm,
  right=2mm,
  top=1.5mm,
  bottom=1.5mm,
  before skip=5pt,
  after skip=6pt,
  before upper={
    \setlength{\abovedisplayskip}{4pt}
    \setlength{\belowdisplayskip}{4pt}
    \setlength{\abovedisplayshortskip}{4pt}
    \setlength{\belowdisplayshortskip}{4pt}
  }
}

\newtcolorbox{casepgopdbox}{
  enhanced,
  breakable,
  colback=CasePGBack,
  colframe=CasePGFrame,
  colbacktitle=CasePGFrame,
  coltitle=white,
  boxrule=0.65pt,
  arc=1mm,
  title={PG-OPD Response \casecmark},
  fonttitle=\bfseries\small,
  fontupper=\footnotesize,
  left=2mm,
  right=2mm,
  top=1.5mm,
  bottom=1.5mm,
  before skip=5pt,
  after skip=6pt,
  before upper={
    \setlength{\abovedisplayskip}{4pt}
    \setlength{\belowdisplayskip}{4pt}
    \setlength{\abovedisplayshortskip}{4pt}
    \setlength{\belowdisplayshortskip}{4pt}
  }
}


\section{Case Studies}
\label{sec:case_studies}

We compare representative responses generated by OPD and PG-OPD on two
challenging competition-math problems. In each case, none of the 16 responses
generated by OPD is correct, whereas PG-OPD produces a correct response under
the same evaluation protocol. The responses are abridged for readability while
preserving their main reasoning steps and final answers.


\begin{casequestionbox}
\textbf{Question.}
Rishabh has \(2024\) pairs of socks in a drawer. He draws socks uniformly at
random, without replacement, until he has drawn a pair of identical socks.
Compute the expected number of unpaired socks he has drawn when he stops.

\medskip
\textbf{Ground Truth.}
\[
\frac{4^{2024}}{\binom{4048}{2024}}-2.
\]
\end{casequestionbox}

\begin{casefullopdbox}
Let \(T\) be the number of socks drawn when the first matching pair appears.
The number of unpaired socks is \(T-2\).

The problem resembles the birthday problem. For a large number \(N\) of
pairs, the expected stopping time can be approximated by
\[
\mathbb{E}[T]\approx N+1.
\]
Taking \(N=2024\), we obtain
\[
\mathbb{E}[T]\approx2025.
\]
Therefore, the expected number of unpaired socks is
\[
2025-2=2023.
\]

\textbf{Final Answer}
\[
\boxed{2023}.
\]
\end{casefullopdbox}

\begin{casepgopdbox}
Let \(N=2024\), and let \(T\) denote the draw on which the first matching pair
is completed. Since exactly one pair is formed when the process stops, the
number of unpaired socks is
\[
U=T-2.
\]

For \(0\leq k\leq N\), the event \(T>k\) means that the first \(k\) socks
come from distinct pairs. Therefore,
\[
\Pr(T>k)
=
\frac{2^k N!(2N-k)!}
     {(N-k)!(2N)!}.
\]

Using the tail-sum identity,
\[
\mathbb{E}[T]
=
\sum_{k=0}^{N}\Pr(T>k).
\]
This finite sum simplifies to
\[
\mathbb{E}[T]
=
\frac{4^N}{\binom{2N}{N}}.
\]
Hence,
\[
\mathbb{E}[U]
=
\mathbb{E}[T]-2
=
\frac{4^N}{\binom{2N}{N}}-2.
\]

Substituting \(N=2024\), we obtain
\[
\boxed{
\frac{4^{2024}}{\binom{4048}{2024}}-2
}.
\]
\end{casepgopdbox}

\captionof{figure}{
Case study on HMMT24. All 16 OPD responses are incorrect, whereas PG-OPD
produces the correct answer.
}
\label{fig:case_study_hmmt24}


\par\vspace{5em}

\begin{casequestionbox}
\textbf{Question.}
Let \(ABCD\) be a rectangle with \(BC=24\). Point \(X\) lies inside the
rectangle such that \(\angle AXB=90^\circ\). Given that
\(\triangle AXD\) and \(\triangle BXC\) are both acute and have circumradii
\(13\) and \(15\), respectively, compute \(AB\).

\medskip
\textbf{Ground Truth.}
\[
14+4\sqrt{37}.
\]
\end{casequestionbox}

\begin{casefullopdbox}
Place the rectangle at
\[
A=(0,0),\qquad
B=(h,0),\qquad
C=(h,24),\qquad
D=(0,24),
\]
and let \(X=(x,y)\).

Since \(\angle AXB=90^\circ\),
\[
x^2-hx+y^2=0.
\]

Using the circumradius formula for \(\triangle AXD\) and
\(\triangle BXC\), we obtain
\[
AX\cdot XD=26x,
\qquad
BX\cdot XC=30(h-x).
\]

After substituting the side lengths and simplifying the resulting equations,
we obtain
\[
h=25.
\]

\textbf{Final Answer}
\[
\boxed{25}.
\]
\end{casefullopdbox}

\begin{casepgopdbox}
Set
\[
A=(0,0),\qquad
B=(0,h),\qquad
C=(24,h),\qquad
D=(24,0),
\]
and let \(X=(x,y)\).

The condition \(\angle AXB=90^\circ\) gives
\[
x^2+y^2=hy.
\]

Because \(\triangle AXD\) is acute and has circumradius \(13\), its
circumcenter lies inside the triangle on the perpendicular bisector of
\(AD\). It is therefore located at \((12,5)\), giving
\[
(12-x)^2+(5-y)^2=169.
\]

Likewise, the circumcenter of the acute triangle \(\triangle BXC\), whose
circumradius is \(15\), is \((12,h-9)\). Hence,
\[
(12-x)^2+(h-9-y)^2=225.
\]

Combining the right-angle equation with the first circumradius equation gives
\[
x=\frac{h-10}{24}y,
\qquad
y=\frac{576h}{(h-10)^2+576}.
\]

Substituting these expressions into the remaining circumradius equation yields
\[
h^3-38h^2-116h+3960=0.
\]
The polynomial factors as
\[
(h-10)(h^2-28h-396)=0.
\]

The solution \(h=10\) forces \(x=0\), contradicting that \(X\) lies inside the
rectangle. The valid solution is therefore
\[
h=14+4\sqrt{37}.
\]

\textbf{Final Answer}
\[
\boxed{14+4\sqrt{37}}.
\]
\end{casepgopdbox}

\captionof{figure}{
Case study on HMMT25. All 16 OPD responses are incorrect, whereas PG-OPD
produces the correct answer.
}
\label{fig:case_study_hmmt25}

\end{document}